\documentclass[a4paper,fleqn]{cas-dc}

\usepackage[authoryear]{natbib}

\usepackage{amsmath,amsfonts}
\usepackage{booktabs}
\usepackage{hyperref}
\usepackage{amsfonts}
\usepackage{xcolor}
\usepackage{color}
\usepackage{amsthm}
\definecolor{mygray}{gray}{.9}
\usepackage{colortbl}

\def\tsc#1{\csdef{#1}{\textsc{\lowercase{#1}}\xspace}}
\tsc{WGM}
\tsc{QE}

\begin{document}
\let\WriteBookmarks\relax
\def\floatpagepagefraction{1}
\def\textpagefraction{.001}

\shorttitle{Multi-Tailed Vision Transformer for Efficient Inference}    
\shortauthors{Y. Wang, B. Du, W. Wang et al.}  
\title [mode = title]{Multi-Tailed Vision Transformer for Efficient Inference}  



%

\author[1]{Yunke Wang}[orcid=0009-0003-9796-530X]
\ead{yunke.wang@whu.edu.cn}

\author[1]{Bo Du}[orcid=0000-0002-0059-8458]
\ead{dubo@whu.edu.cn}

\author[2]{Wenyuan Wang}[orcid=0009-0007-5287-1296]
\ead{wenyuan.wang@whu.edu.cn}

\author[3]{Chang Xu}[orcid=0000-0002-4756-0609]
\ead{c.xu@sydney.edu.au}

\affiliation[1]{organization={School of Computer Science, Wuhan University},
            city={Wuhan},
            postcode={430072}, 
            country={China}}

\affiliation[2]{organization={School of Electric Information, Wuhan University},
            city={Wuhan},
            postcode={430072}, 
            country={China}}

\affiliation[3]{organization={School of Computer Science, The University of Sydney},
            city={Sydney},
            country={Australia}}




\begin{abstract}
Recently, Vision Transformer (ViT) has achieved promising performance in image recognition and gradually serves as a powerful backbone in various vision tasks. To satisfy the sequential input of Transformer, the tail of ViT first splits each image into a sequence of visual tokens with a fixed length. Then, the following self-attention layers construct the global relationship between tokens to produce useful representation for the downstream tasks. Empirically, representing the image with more tokens leads to better performance, yet the quadratic computational complexity of self-attention layer to the number of tokens could seriously influence the efficiency of ViT's inference. For computational reduction, a few pruning methods progressively prune uninformative tokens in the Transformer encoder, while leaving the number of tokens before the Transformer untouched. In fact, fewer tokens as the input for the Transformer encoder can directly reduce the following computational cost. In this spirit, we propose a Multi-Tailed Vision Transformer (MT-ViT) in the paper. MT-ViT adopts multiple tails to produce visual sequences of different lengths for the following Transformer encoder. A tail predictor is introduced to decide which tail is the most efficient for the image to produce accurate prediction. Both modules are optimized in an end-to-end fashion, with the Gumbel-Softmax trick. Experiments on ImageNet-1K demonstrate that MT-ViT can achieve a significant reduction on FLOPs with no degradation of the accuracy and outperform compared methods in both accuracy and FLOPs.
\end{abstract}


\maketitle

\section{Introduction}
The great success of Transformer~\citep{vaswani2017attention, devlin2018bert, brown2020language} in Natural Language Processing (NLP) has drawn computer vision researchers' attention. There have been some attempts on adopting the Transformer as an alternative deep neural architecture in computer vision. Vision Transformer (ViT)~\citep{dosovitskiy2020image} is a seminal work that employs a fully Transformer architecture to address the image classification task. By first splitting an image into multiple local patches, ViT can then form a visual sequence for the Transformer input. The self-attention mechanism in ViT is capable of measuring the relationship between any two local patches and then information of patches are aggregated to produce a high-level representation for the image recognition task. 

Following ViT, a number of of variants \citep{yuan2021tokens, han2021Transformer, pan2021scalable, touvron2021training, jiang2021all,zhou2021deepvit,guo2022cmt, wang2021crossformer} have been developed. For example,  DeiT \citep{touvron2021training}, without the pre-training on an extra large-scale dataset (\textit{e.g.},  JFT-300M \citep{sun2017revisiting}), for the first time boosts ViT to achieve the state-of-the-art performance on the ImageNet-1K \citep{deng2009imagenet} benchmark. T2T-ViT \citep{yuan2021tokens}, which can also be trained from scratch on the ImageNet-1K benchmark, proposes to boost the exchange of local information and global information with a T2T-module before the transformer encoder. CrossViT \citep{chen2021crossvit}  exploits multi-scale features of the image in vision transformer and TNT \citep{han2021Transformer} focuses on investigating the attention inside a single patch and proposes the divide of a single patch into multiple smaller patches. CrossFormer~\citep{wang2021crossformer} proposes a novel method that utilizes patches of different sizes to construct cross-scale attention, which shows great improvement in several important vision benchmarks.
These efforts have thus ensured vision transformer to be a strong substitution (\textit{i.e.}, Swin Transformer \citep{liu2021swin} and Twins \citep{chu2021twins}) for CNNs architecture \citep{krizhevsky2012imagenet, he2016deep, tan2019efficientnet} in vision tasks. However, compared with CNNs, vision transformers do not show a significant decrease in the computational cost, or sometimes consume even more. 

As the computational cost of the transformer is quadratic to the sequence length, a natural idea is therefore to decrease the number of tokens in the transformer for a potential acceleration. But the number of tokens is also a key factor to the accuracy, which thus requests a non-trivial strategy of screening tokens to achieve a trade-off between accuracy and computational cost. Inspired by the pruning in CNNs \citep{he2017channel}, a few works suggest pruning tokens in vision transformers for efficient inference. PoWER-BERT \citep{goyal2020power} observes that the redundancy in Transformer gradually grows from the shallow layer to the high layer, and the token redundancy can be measured through their attention scores. This insightful observation then motivated pruning methods that progressively prune tokens in vision transformers \citep{rao2021dynamicvit, tang2022patch, chen2021chasing, yin2022vit, xu2022evo}.
\begin{figure}[!t]
	\centering
	\includegraphics[width=0.48\textwidth]{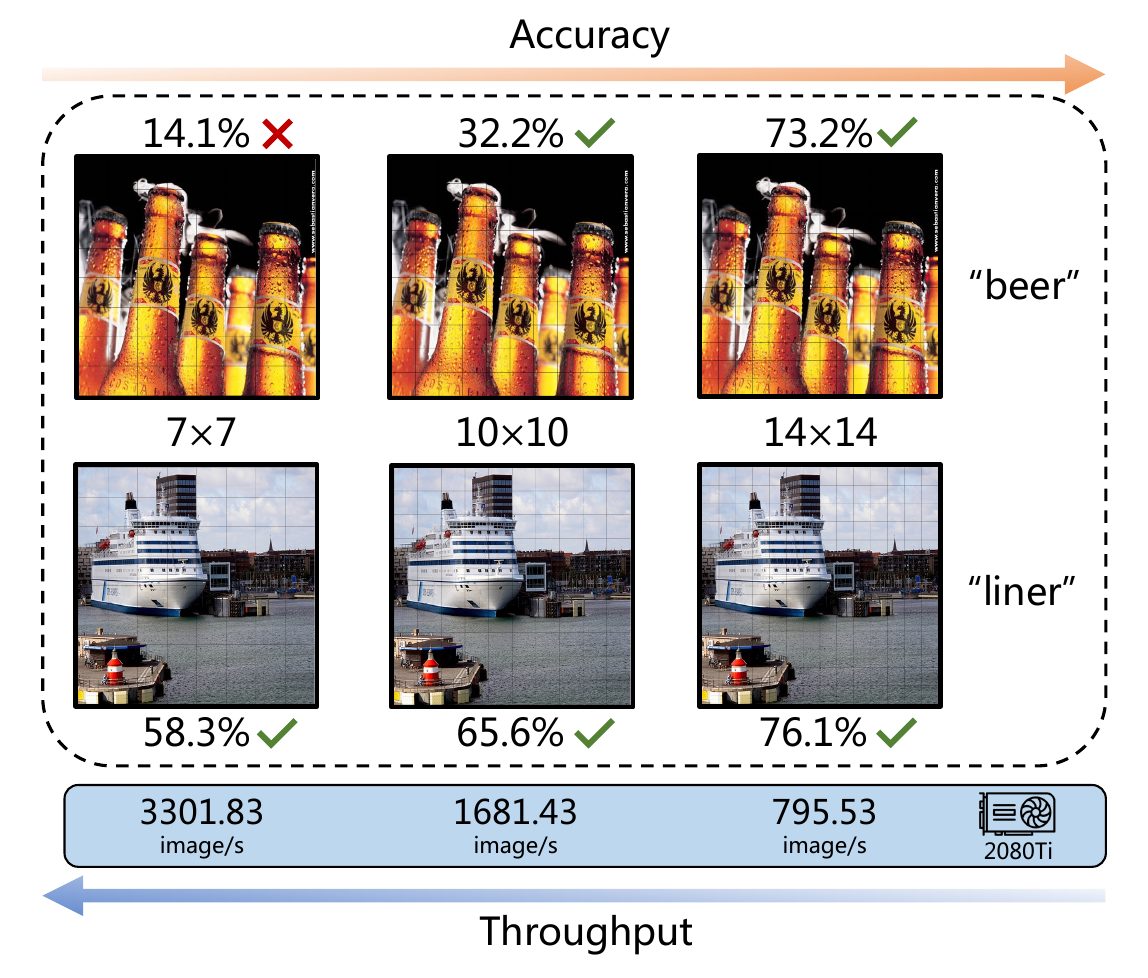}
	\caption{The throughput and confidence of prediction from DeiT-S, with a different number of tokens, \textit{e.g.}, 7$\times$7, 10$\times$10 and 14$\times$14. The `tick' and `cross' sign denote the right and false prediction respectively.}
	\label{fig_1}
\end{figure}

These pruning methods have shown great success in reducing the number of tokens for efficient inference in vision transformers, \textit{i.e.}, all of them can preserve the accuracy under a nearly $40\%$ FLOPs reduction. But their progressively pruning strategy only deals with the Transformer encoder, while leaving the very beginning number of tokens at the input of vision transformers untouched. In fact, the upper bound of the overall computational cost is mainly determined by the number of tokens in the ``image-to-tokens'' step before the Transformer encoder. If we split an image into fewer patches, there will be a shorter visual sequence input, which implies a higher inference speed by the following vision transformer. Also, the different complexities of images will favor a customized number of patches to be split. An easy image could also be accurately recognized with fewer patches, while difficult images could need a more fine-grained split to guarantee the recognition accuracy, as shown in Figure \ref{fig_1}.

This paper introduces the Multi-Tailed Vision Transformer (MT-ViT), a novel approach that optimizes the number of patches used to represent input images, thereby reducing computational cost. Unlike traditional Vision Transformers, which typically employ a single "tail" in their "image-to-tokens" module, MT-ViT incorporates multiple tails that can produce visual sequences of varying lengths. By projecting patches of corresponding resolution into the same d-dimensional token, we enable sharing of the subsequent Transformer encoder.
When processing an input image, a tail predictor is trained to determine which tail should be used to generate the visual sequence. As tail selection is a non-differentiable process, we employ the Gumbel-Softmax technique to optimize MT-ViT and the tail predictor in an end-to-end fashion. We evaluate MT-ViT on both small scale datasets (\textit{e.g.}, CIFAR100 \citep{krizhevsky2009learning}, TinyImageNet \citep{chrabaszcz2017downsampled}) and large scale datasets (\textit{e.g.}, ImageNet-1K \citep{deng2009imagenet}) on top of various backbone (\textit{e.g.}, DeiT \citep{touvron2021training}, T2T-ViT \citep{yuan2021tokens} and MiniViT \citep{zhang2022minivit}). We highlight the contributions of MT-ViT as follows.
\begin{itemize}
    \item We propose a novel approach that optimizes the number of patches used to represent input images, thereby reducing the computational cost of the vision transformer. Our approach servers as a general architecture for efficient inference of vision transformer and can integrate with various ViT backbones.
    \item Empirical results demonstrate MT-ViT's ability to maintain accuracy while achieving an up to 70\% reduction in FLOPs on CIFAR100 and TinyImageNet, and achieve an obvious advantage over comparison pruning methods on ImageNet-1K benchmark.
    \item Visualizations of the tail predictor's results reveal its capacity to automatically translate the difficulty of the image. Its decision is basically consistent with human's visual judgement. 
\end{itemize}

\section{Related Work}
In this section, we discuss the development of vision transformer first. Then. we introduce some efficient inference methods for vision transformer and neural architecture search methods in transformer.

\subsection{Vision Transformer}
The transformer has been widely used in NLP community \citep{vaswani2017attention,devlin2018bert, brown2020language} and achieved great success. Inspired by this major success of transformer architectures in the field of NLP, researchers have recently applied transformer to computer vision (CV) tasks \citep{han2020survey, qiu2022ivt, he2021transreid, qiu2022learning, jiao2023dilateformer,zhao2022spatial, rendon2023crimenet, han2022dual,chopin2023interaction,jia2022learning,chen2023transformer, gao2023generalized,jahanbakht2022sediment}. The early attempts about applying Transformer to vision tasks focus on combining convolution with self-attention.
DETR \citep{carion2020end} is proposed for object detection task, which first exploits the convolution layer to extract visual features and then refines features with Transformer. BotNet \citep{srinivas2021bottleneck} replaces the convolution layers with multi-head self-attention layer at the last stage of ResNet \citep{he2016deep} and achieves good performance.

ViT \citep{dosovitskiy2020image} is the first work to introduce a fully Transformer architecture directly into vision tasks. By pre-training on massive datasets like JFT-300M \citep{sun2017revisiting} and ImageNet-21K \citep{deng2009imagenet}, ViT achieves state-of-the-art performance on various image recognition benchmarks. However, ViT's performance is relatively modest when trained on mid-sized datasets such as ImageNet-1K \citep{deng2009imagenet}. In comparison to ResNet \citep{he2016deep} of similar size, ViT obtains slightly lower accuracy. The primary reason for this discrepancy is that transformers lack certain inductive biases about images such as locality and translation equivariance. These biases are critical for generalization, particularly when training vision transformers with limited training data. Later, DeiT \citep{touvron2021training} manages to solve this data-efficient problem by simply modifying the Transformer and proposing a Knowledge Distillation (KD) \citep{hinton2015distilling} optimization strategy, with improved accuracy in ImageNet-1K. Some following works \citep{yuan2021tokens,han2021Transformer, chen2021crossvit} focus on exploiting the local information of the image, which lead to a significantly increasing performance. Also, some works \citep{liu2021swin,pan2021scalable,lee2022mpvit} try to adopt a deep-narrow structure like CNNs to produce multi-scale features for the downstream intensive prediction task. Since the vision transformer is known to suffer from a huge number of parameters, there are also some attempts to achieve parameter reduction while retaining the same performance \citep{mehta2021mobilevit, zhang2022minivit, wu2022tinyvit}. 
At present, vision transformer has been applied into various vision tasks, such as medical image analysis~\citep{grigas2023improving,xu2022lssanet}, NeRF~\citep{chen2023interactive}, supersampling~\citep{guo2023ultra}, and multi-information fusion~\citep{xu2023multi,odusami2023pixel,zhang2023vit}.


\subsection{Efficient Transformer}
Despite ViT's impressive performance in vision tasks, achieving good performance requires significant computational resources. Consequently, researchers are interested in developing a more efficient Transformer architecture~\citep{fournier2023practical, chitty2023survey, kim2023full}. Network pruning and compression have been widely used in CNNs to speed up neural network inference (\textit{i.e.}, filter pruning \citep{he2017channel, liu2017learning, tang2020scop,chitty2020calibration}) and adopt light-weight nerual network architecture~\citep{wang2018towards,wang2018learning}. Some researchers \citep{goyal2020power, rao2021dynamicvit, tang2022patch, su2022vitas, chen2021chasing, pan2021ia, xu2022evo, yin2022vit, liang2022not} have been inspired by this idea and have attempted to use token pruning to identify and eliminate inferior tokens in order to improve efficiency.
More recently, A-ViT \citep{yin2022vit} achieves efficient computing by adaptively halting tokens that are deemed irrelevant to the task, thereby enabling dense computation only on the active informative tokens. This module reuses existing block parameters and utilizes a single neuron from the last dense layer in each block to compute the halting probability, requiring no additional parameters or computations. EViT \citep{liang2022not} highlights the importance of class tokens and images by their attention scores. ATS \citep{fayyaz2022adaptive} introduces a parameter-free module that scores and adaptively samples significant tokens. Above pruning methods are mostly based on scores, which only keep tokens with highest score. However, this selecting scheme can cause redundancy and information loss. Token Pooling~\citep{marin2023token} takes a different path to downsample tokens. It forms multiple clusters to approximate the set of tokens, then selects the cluster centers. Thus, the output tokens are a more accurate representation of the original token set than the score-based methods.

Some adaptive methods aim to reduce the inference time of the model conditionally. Motivated by this idea, \citep{bakhtiarnia2022single} proposes several multi-exit architectures for dynamic inference in vision transformers. The core idea is to conduct an early exit in the middle layer when the prediction confidence is above the threshold. Dynamic-Vision-Transformer (DVT) \citep{wang2021not} reduces the computational cost by considering cascading three Transformers with increasing numbers of tokens, which are sequentially activated in an adaptive fashion during the inference time. Specifically, an image is first sent into the Transformer with a fewer number of tokens. By investigating the prediction confidence, DVT decides whether to proceed to use the next Transformer. The subsequent model also considers utilizing the intermediate feature of the former Transformer. 

\subsection{Neural Architecture Search}
Neural Architecture Search (NAS)~\citep{chitty2023neural} is designed to automatically create neural architectures for networks. Early NAS methods were computationally intensive, requiring the training and evaluation of a large number of architectures. However, differentiable NAS approaches like DARTS~\citep{liu2018darts}, DNAS~\citep{wu2019fbnet}, and ProxylessNAS~\citep{cai2018proxylessnas} have emerged, enabling joint and differentiable optimization of model weights and architecture parameters through gradient descent. This significantly reduces computational costs.
Recently, vision transformer has draw considerable attention in the research community and some researchers are exploring the use of neural architecture search (NAS)~\citep{chitty2022neural,chitty2022neural1} to find an efficient ViT architecture. For example, AutoFormer~\citep{chen2021autoformer} combines the weights of various blocks in the same layers during supernet training. NASViT~\citep{gong2021nasvit} aims to alleviate the gradient conflict issue in NAS and ensure the efficiency of the ViT model. BossNAS~\citep{li2021bossnas} implemented the search with an self-supervised training scheme and leveraged a hybrid CNN-transformer search space for boosting the performance.

Our paper introduces a differentiable training approach for the tail predictor, which bears similarities to NAS training. However, our paper's objective differs, as we do not seek a specific structure for the Vision Transformer. Instead, our focus is on enhancing the efficiency of the Vision Transformer by dynamically allocating different samples with varying computational budgets, which is achieved through the use of dynamic networks.

\section{Preliminaries}
\label{sec3}
A standard Transformer in NLP tasks normally requires a 1D sequence of token embedding as the input. To handle 2D images, the tail of ViT splits an image $X\in \mathbb{R}^{h\times w\times c}$ into $N$ independent patches $X_p$ in a $p\times p$ resolution and then projects each local patch $X_p$ into a $d$ dimensions embedding to form a visual sequence $Z_p \in \mathbb{R}^{N\times d}$.
Similar to BERT \citep{devlin2018bert}, ViT also introduces a learnable class token $Z_{cls}$ into the input sequence (\textit{i.e.}, $Z=[Z_{cls}, Z_p]$ ). Subsequently, ViT learns a representation of the image with the following $L$-layers Transformer encoder, which mainly consists of two alternating components (\textit{i.e.}, Multi-head Self-Attention (MSA) module and Multi-Layer Perceptron (MLP) module). 
\subsection{Multi-head Self-Attention}
MSA module calculates the relationship between any two tokens to generate the attention map $A$ with self-attention layer. Given a sequential input $Z \in \mathbb{R}^{(N+1)\times d}$, standard self-attention module first projects $Z$ linearly into three embedding called query $\textbf{Q}$, key $\textbf{K}$ and value $\textbf{V}$ respectively,
\begin{align}
    [\textbf{Q}, \textbf{K}, \textbf{V}] = ZW_{qkv},
\end{align}
where $W_{qkv}=[W_q, W_k, W_v]\in \mathbb{R}^{d\times3d}$ denotes the linear projection operator. Then, the attention map $A$ is calculated by the dot operation between query $\textbf{Q}$ and key $\textbf{K}$ and then the attention map $A$ is finally applied to weight the value embedding $\textbf{V}$. The whole process of self-attention can be defined as follows,
\begin{align}
    {\rm SA}(\mathbf{Q,K,V})=A\textbf{V}={\rm softmax}(\frac{\textbf{Q}\textbf{K}^\mathrm{T}}{\sqrt{d}})\textbf{V}.
\end{align}
Compared to the vanilla self-attention, multi-head self-attention runs $k$ self-attention operations in parallel. The output of MSA can be formulated as follows,
\begin{align}
    {\rm MSA}(\mathbf{Q,K,V})=\rm{Concat}\big[{\rm softmax}\Big(\frac{\textbf{Q}^{h}{\textbf{K}^{h}}^\mathrm{T}}{\sqrt{d}}\big)\textbf{V}^h\Big]^{H}_{h=1}.
\end{align}

\subsection{Multi-Layer Perceptron}
The MLP module is applied after MSA module for representing feature and introducing non-linearity. Denoting $Z'$ as the output of MSA module, MLP can be defined as follows,
\begin{align}
{\rm MLP}(Z')=\phi(Z'W_{fc}^{a})W_{fc}^{b},
\end{align}
where $W_{fc}^{a}$ and $W_{fc}^{b}$ denote the fully-connected layer, and $\phi$ denotes the non-linear activate function (\textit{e.g.}, GELU).

\subsection{Transformer Encoder}
The Transformer encoder is constructed by stacking MSA module and MLP module with residual connection. Therefore, the encoder can be defined as 
\begin{align}
    Z^{\prime}_{l} &= Z_{l-1}+{\rm MSA}({\rm LN}(Z_{l-1})), \\
    Z_{l}&=Z^{\prime}_{l}+{\rm MLP}({\rm LN}(Z^{\prime}_{l})),
\end{align}
where ${\rm LN}(\cdot)$ is the layer normalization for stable the training of Transformer.

\subsection{Analysis of the computation complexity}
The floating-point operations (FLOPs) is commonly used as a metric to measure the theoretical computational cost of the model. After summing up the FLOPs of all the operations in the Transformer encoder, we find that MSA and MLP contribute the most to FLOPs. Specifically, the FLOPs of MSA are $4N{d^2}+2{N^2}d$ and the FLOPs of MLP are $2N{d^2}r$, where $r$ is the dimension expansion ratio of the fully-connected layer in MLP.
From the analysis above, we can observe that the FLOPs of the Transformer encoder is quadratic to the embedding dimension $d$ and number of tokens $N$.
Due to the large number of $d$ and $N$ (usually hundred), the computational cost of vision transformer is large. However, if we reduce these two hyper-parameters for ViT model, we can easily obtain a rapid decrease in computational cost without modifying the architecture of the Transformer.

Adjusting the embedding dimension $d$ has been well considered in various ViT backbones. By setting $d$ from a small value to a large value, we can obtain ViT model with increasing FLOPs (\textit{i.e.}, ViT-Small, ViT-Base and ViT-Large). In return for the high complexity, the larger the model size is, the higher accuracy the model will normally achieve. Another way is to consider the number of tokens $N$, which can be up to the resolution of the local patch. For an image of 224$\times$224 size, ViT model normally splits the image into non-overlap patches with 16$\times$16 size, so the number of tokens $N$ is equal to $(224/16)^2=196$. By setting different patch resolutions, we can obtain different ViT models, \textit{e.g.}, ViT-Base/14, ViT-Base/16 and ViT-Base/32.

\begin{figure*}[!tbp]
    \centering
    \includegraphics[width=1.91\columnwidth]{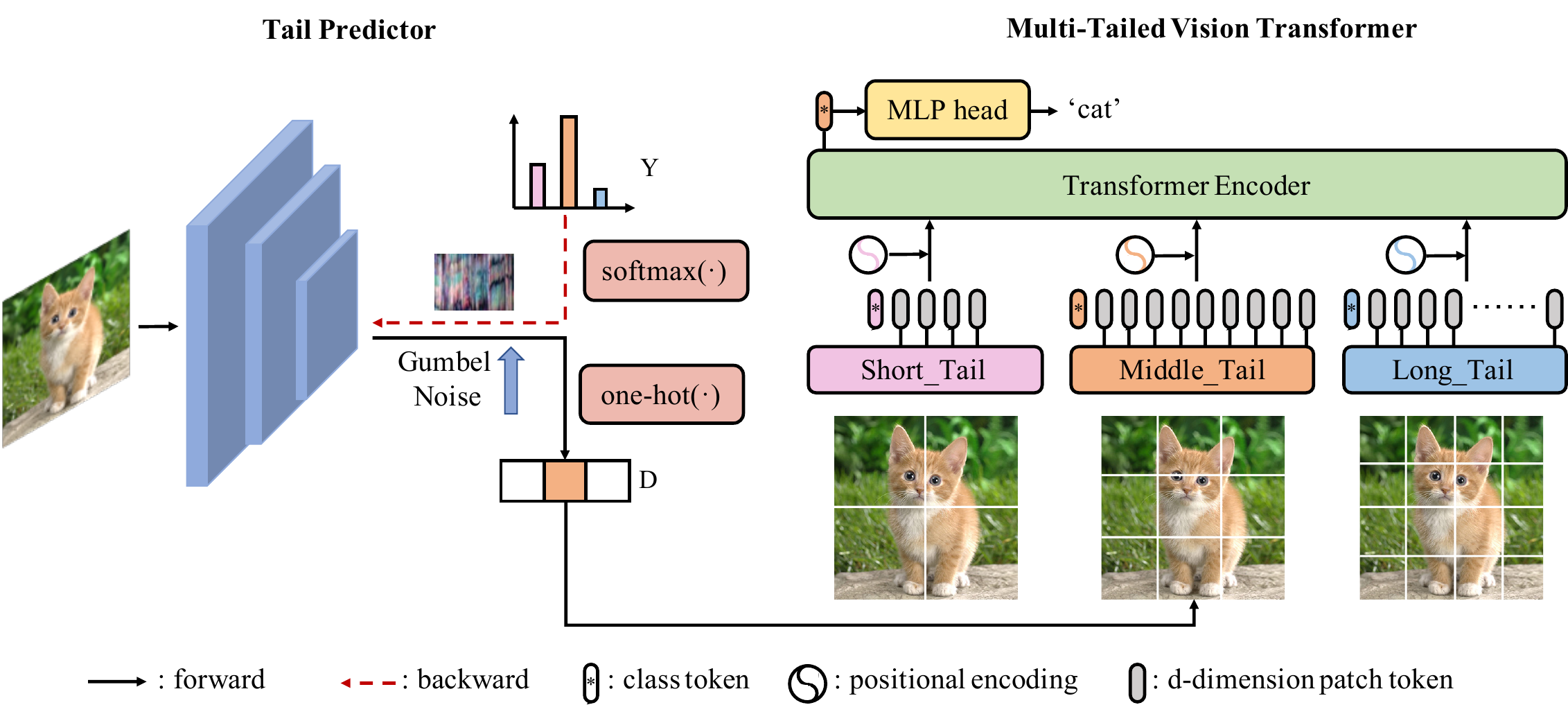}
    \caption{The framework of proposed method contains two main components: scale predictor $\pi_\theta$ and multi-tailed vision transformer (MT-ViT). The number of scale $K$ is set to 3. (a) The tail predictor is a CNN-based model that determines the appropriate tail for the image. (b) By using the multiple tails in MT-ViT, patches with different sizes are all projected into a $d$ dimension embedding. This makes it possible to share the Transformer encoders and MLP head. }
    \label{msvit}
\end{figure*}
\section{Methodology}
While the Vision Transformer demonstrates promising performance in vision tasks, the escalating computational cost for efficient inference has become a focal point for researchers. As discussed in Section \ref{sec3}, decreasing the number of tokens $N$ can be an effective way to ensure a significant reduction on FLOPs. Existing Transformer pruning methods have explored the screening out of uninformative tokens at intermediate layers. Nevertheless, there remains potential for even greater computational cost reduction by reducing the number of tokens at earlier positions. Based on this idea, we further explore reducing tokens at the ``image-to-tokens'' step for a potential FLOPs reduction.

In the ``image-to-tokens'' step before the Transformer, an image $X\in \mathbb{R}^{h\times w\times c}$ is decomposed into a sequence of non-overlap patches with $p\times p$ size and the number of patches $N$ is equal to $(h\times w)/{p^2}$. Therefore, the resolution of the local patch can determine the length of the visual sequence and further affects the accuracy and computational cost of ViT. With a fine-grained patch size, ViT could reach higher performance with increasing FLOPs. On the contrary, with a coarse-grained patch size, the performance of ViT decreases with a reduction of FLOPs. 

Motivated by this idea, we consider leveraging the advantages of both fine-grained patch size and coarse-grained patch size to achieve the trade-off between accuracy and efficiency. Intuitively, easy images could be accurately recognized with a coarse-grained patch size while difficult images often require a fine-grained patch split to achieve an accurate prediction. 
To achieve this, we firstly need a network module that can be compatible with both fine-grained patch and coarse-grained patch. A natural idea is to construct a stack of $K$ independent vision transformers $F(X)=[f_1, f_2, ...., f_K]$, which are pre-trained with its corresponding $K$ patch sizes (\textit{i.e.}, $p_1, p_2,..., p_K$). 
To decide which Transformer is suitable for the image, a one-hot decision vector $D\in [0,1]^K$ is introduced to determine which patch size $p$ is proper for the given image. With an optimal decision $D$, we can minimize computational cost as far as possible while preserving the accuracy of the model. 

The basic workflow is illustrated as follows. Denoting $m$ as the number of classes and $f_k(X)\in \mathbb{R}^{m}$ as the logit of the $k$-th stacked ViT model, the final prediction of the whole stacked model $\widehat{F}$ is the sum of each Transformer with weight $D$, which can be formulated as
\begin{align}
\widehat{F}(X)=D \odot F(X)=\sum_{i=0}^K D_i f_i(X).
\label{fprediction}
\end{align}
The classification loss $L_{\rm cls}$ for image $X$ can be written as
\begin{align}
L_{\rm cls}={\rm Cross\_Entropy}(\widehat{F}(X), Y_{\rm cls}),
\label{floss}
\end{align}
where ${\rm Cross\_Entropy}(\cdot$) is the softmax cross-entropy loss and $Y_{\rm cls}$ is the class label related to $X$.

\subsection{Multi-Tailed Vision Transformer}
In the stacked Transformer framework, a stack of $K$ independent vision transformers is used to process multi-scale visual sequences respectively. But the stacked Transformers directly lead to a $K$ times number of parameters than that of a single vision transformer model, which is horribly parameter-inefficient. Normally, setting some shared parameters and layers among these independent Transformers is essential for reducing the parameter of the model when processing multi-scale input.

Inspired by this idea, we adopt a shared Transformer encoder in the stacked Transformer $F$, which results in the multi-tailed vision transformer (MT-ViT). The ``image-to-tokens'' step is referred as the tail of ViT. MT-ViT adopts $K$ independent tails for projecting patches in different sizes into a $d$-dimension embedding, and then shares all parameters of Transformer encoders and classification head, as shown in Figure \ref{msvit}. The instance-aware tails are conditioned to images and projects patches of different sizes into vectors of the same dimension to satisfy the input requirement of the public Transformer encoder. As the inside architecture of Transformer encoder has not been changed, we can make MT-ViT serve as a general backbone that is compatible with the mainstream ViT backbones.

MT-ViT adopts multiple tails before the Transformer encoder. For the $i$-th tail, it should receive patches $x_p$ of size $p_i$ and then proceed $x_p$ into the projection function $\mathcal{T}_i(\cdot)$ to get the embedding $z_i$, where $z_i=\mathcal{T}_i(x_p)\in \mathbb{R}^d$.
For different vision transformer backbones, the projection function $\mathcal{T}$ can be quite different, therefore we should re-design the tail to produce dynamic sequences of $d$-dimension embedding when MT-ViT is equipped with different ViT backbones. 

\subsection{Dynamic Tail Selection}
Optimizing MT-ViT (Eq. (\ref{floss})) could be intractable for the following reason. In the classical image classification task, images and their corresponding label are the only available information in the training set, while the optimal tail selection $D$ remains unknown. We therefore expect an appropriate estimation on decision $D$, so that we can well indicate the ``easiness'' of the image and make full use of the multi-tailed vision transformer backbone. To solve this problem, we propose a CNN-based tail predictor $\pi_\theta$ to automatically distinguish the ``easiness'' of the image and output the proper decision $D$. Given the image $X$, the policy $\pi_\theta$ outputs the categorical distribution,
\begin{align}
\zeta=[\zeta^{(1)},\zeta^{(2)},...,\zeta^{(K)}]=\pi_\theta(X),
\end{align}
where $\zeta^{(i)}$ represents the probability of choosing the $i$-th patch size. To generate decision $D$, we need to sample from $\zeta$ or select the highest probability of $\zeta$ to get the one-hot vector. However, both sampling from softmax distribution and selecting the highest value in $\zeta$ are non-differentiable, which precludes the back-propagation and impedes the end-to-end training. 

\subsubsection{Differentiable Tail Sampling}
Gumbel-Softmax trick~\citep{jang2016categorical} is a technique that can be used to train models with discrete latent variables through backpropagation. It introduces a continuous relaxation of discrete random variables, allowing for the use of gradient-based optimization methods. Therefore, we apply the Gumbel-Softmax trick to ensure a differentiable tail sampling during the forward propagation.
First, we form a Gumbel-Softmax distribution and transform the soft probability $\zeta$ to a discrete variable $D\in \{0,1\}^K$ as follows,
\begin{align}
D={\rm one\_hot}\Big(\mathop{\arg\max}\limits_{i}[g^{(i)}+\log {\zeta}^{(i)}]\Big)
 \label{gumbel}
\end{align}
where $g^{(1)},...,g^{(k)}$ are \textit{i.i.d} random noise samples drawn from Gumbel(0, 1) distribution. The Gumbel distribution has shown to be stable under $\max$ operations \citep{maddison2016concrete}.
By applying the inverse-CDF transformation, $g^{(i)}$ can be computed as
\begin{align}
    g^{(i)}=-\log(-\log(u^{(i)})), \quad  u^{(i)}\sim U(0,1),
\end{align}
where $U(0,1)$ is the Uniform distribution. Since the $\arg\max(\cdot)$ in Eq. (\ref{gumbel}) is still a non-differentiable operation, we further use the softmax function as a continuous, differentiable approximation to replace $\arg\max(\cdot)$. A differentiable sampling $Y=[y^{(1)},y^{(2)},...,y^{(K)}] \in \{0,1\}^K$ can be written as,
\begin{align}
    y^{(i)}=\frac{\exp(\log{\zeta^{(i)}+g^{(i)})/\tau}}{\sum_{i=1}^{m}\exp[(\log{\zeta^{(i)}+g^{(i)})/\tau]}}, \quad i=0,1,...,K,
\end{align}
where $\tau \in (0, +\infty)$ is the temperature parameter to adjust the Gumbel-Softmax distribution. $Y$ can be regarded as continuous relaxations of one-hot vectors $D$. As the softmax temperature $\tau$ approaches 0, samples from the Gumbel-Softmax distribution approximate one-hot vectors. The sampling of the approximated one-hot vector is refactored into a deterministic function-componentwise addition followed by $\arg\max$ of the parameters $\log {\zeta}$ and fixed Gumbel distribution $-\log(-\log(u))$. The non-differentiable part is therefore transferred to the samples $g$ from the Gumbel distribution.  
This reparameterization trick allows gradients to flow from decision $D$ to the $\theta$ and both tail predictor and MT-ViT can be optimized as a whole with Eq. (\ref{floss}). 

However in our setting, we are constrained to sample discrete values strictly because we can only choose one tail of MT-ViT to make the prediction. Inspired by Straight-Through Gumbel Estimator, in the forward propagation, we discretize $D$ using $\arg\max(\cdot)$ in Eq. (\ref{gumbel}) but use the continuous approximation in the backward propagation by approximating $\nabla_\theta D \approx \nabla_\theta Y$.

\subsubsection{Optimization}
The basic loss function (Eq. (\ref{floss})) defines a softmax cross-entropy loss between the prediction of MT-ViT and the ground-truth label. The multi-tailed vision transformer backbone and the tail predictor are optimized jointly with the Gumbel-Softmax trick. However, Eq. (\ref{floss}) only considers achieving better accuracy, which makes the tail predictor always encourage MT-ViT to activate the tail with the highest accuracy and therefore lead to the collapse of the predictor's training. To solve this problem, we consider adding a FLOPs constraint regularization on the choice of different tails. The total loss can be written as follows,
\begin{align}
    L_{\rm total}=L_{cls} + \lambda L_{f}
\end{align}
where $L_{f}$ is the FLOPs regularization and $\lambda$ is the hyper-parameter to achieve the trade-off between accuracy and FLOPs. The FLOPs regularization can punish the situation when the predictor selects the tail with high computational cost. On the contrary, in those situations where the predictor selects the tail with low FLOPs, the regularization should not contribute any penalty to the total loss. In this spirit, our designed FLOPs constraint regularization is formulated as follows
\begin{align}
    L_{f}=\max\Big(\alpha, \sum_{i=0}^K D_i c_i \Big)-\alpha,
    \label{lossf}
\end{align}
where $c_i$ is the normalized FLOPs of the $i$-th branch and $\alpha$ is the hyper-parameter to adjust the threshold of penalty. While the FLOPs of the selected model is larger than the threshold $\alpha$, $L_{f}$ will punish the prediction. 

In summary, while the first cross-entropy loss encourages the predictor to choose the tail that leads to high accuracy, the second FLOPs regularizer will penalize the situation when the predictor chooses the tail with high FLOPs. Hence, the mechanism of MT-ViT can maintain the accuracy while reducing the computational cost at most.

\subsection{Discussion}
In this subsection, we discuss the difference of MT-ViT with several close-related works (\textit{e.g.}, Multi-scale patch methods and token-pruning methods).
\subsubsection{Compare with multi-scale patch methods}
The idea of multi-scale patch has been investigated in some related works, such as CrossViT~\citep{chen2021crossvit} and MPViT~\citep{lee2022mpvit}. 
CrossViT and MPViT are two great works that utilize the idea of multi-scale patch to design a network backbone. By conducting multi-scale feature aggregation in the transformer layer, they enhance the performance of visual recognition, yet introduce more computational cost. Despite the idea of multi-scale patch embedding also appears in MT-ViT, there are still several distinctive aspects that make MT-ViT different. (1) MT-ViT achieves efficient inference through multi-scale patches, while in CrossViT and MPViT, multi-scale patches primarily contribute to generating multi-scale features,  and the aggregation of multi-scale features can further enhance the performance of visual recognition. (2) CrossViT and MPViT employ fixed networks, activating all branches for various images. MT-ViT introduces a dynamic approach, activating only one tail per image, each corresponding to different computational budgets. The tail predictor dynamically selects the appropriate tail, resulting in a dynamic neural network. (3) CrossViT and MPViT utilize specially designed and fixed networks. In contrast, MT-ViT offers flexibility by allowing the switch to different Vision Transformer backbones, providing adaptability and versatility in the model architecture.

The idea of multi-scale tokens has also been used in DVT~\citep{wang2021not} and we will discuss the empirical results between DVT and MT-ViT in the experiment.

\subsubsection{Compare with token-pruning methods}
Token-pruning methods aim to progressively prune uninformative tokens in the transformer layer and accelerate the inference of Transformer, since the computational cost of Transformer is quadric to the number of tokens. The difference between token-pruning methods and MT-ViT can be summarized in the following aspects.
First, pruning-based methods normally start to reduce the number of tokens at the middle layer. It could still be possible for a greater reduction of computational cost by reducing the number of tokens at the earlier position. Based on this idea, our method reduces tokens by adjusting the patch size in different tails at the very beginning before the transformer encoder. Empirical results also demonstrate the effectiveness of MT-ViT over pruning methods.
Second, pruning methods normally set the predictor/gate module inside the transformer architecture, while the predictor in MT-ViT is placed at the beginning of the transformer. This design could keep the original architecture of transformer as a whole and therefore makes it easier to switch from different backbones.

The pros of token-pruning methods lie in its training cost compared to MT-ViT. Token pruning methods can be conducted on existing ViT backbones. Since the trained ViT model is easy to acquire, token pruning methods do not require to train the model from scratch. Normally, they only need to jointly finetune the transformer backbone and token pruning modules for 30 epochs. However, multi-tailed approach added two tails before the transformer encoder. It requires 300 epochs pre-training from scratch to ensure a satisfactory performance for all tails.

\section{Experiment}
In this section, we conduct extensive experimental analysis on the small-scale datasets, CIFAR100 \citep{krizhevsky2009learning} and TinyImageNet \citep{chrabaszcz2017downsampled}, and large-scale dataset, ImageNet-1K benchmark \citep{deng2009imagenet} to show the performance of our proposed Multi-Tailed Vision Transformer in various aspects. Additionally, we evaluate the performance of MT-ViT on other vision tasks like object detection.
\subsection{Experiment Setting}
\subsubsection{Datasets}
CIFAR100 \citep{krizhevsky2009learning} is a widely-used dataset for image recognition tasks, which contains 50,000 training images and 10,000 test images. There are 100 classes in this dataset, grouped into 20 superclasses. Each image has a ``fine label'' as the class label and a ``coarse label'' as the superclass. In our experiment, we only make use of fine labels. TinyImageNet \citep{chrabaszcz2017downsampled} is a subset of the well-known ImageNet-1K benchmark, which contains 100,000 images with 64$\times$64 resolution with 100 categories. 
ImageNet-1K \citep{deng2009imagenet} is the most popular benchmark to evaluate the classification performance of the deep learning model. It contains 1.28 million training images and 50,000 validation images with 1000 categories. Details of these three datasets are summarized in Table \ref{dataset}.
\begin{table}[!h]
    \centering
    \normalsize
    \caption{Detailed information of datasets used for training.}
    \setlength{\tabcolsep}{1.5mm}{
    \begin{tabular}{|l|c|c|c|c|}
    \hline
   \rowcolor{mygray}
    \textbf{Dataset} & \textbf{Train size} & \textbf{Val size} & \textbf{Classes} & \textbf{Size} \\ \hline \hline
         CIFAR100  & 50,000 & 10,000 & 100 & 32$\times$32   \\
         TinyImageNet  & 100,000 & 10,000 & 100 & 64$\times$64 \\
         ImageNet  & 1,281,167 & 50,000 & 1,000 &  N/A \\
    \hline
    \end{tabular}
    }
    \label{dataset}
\end{table}
\begin{table}[!h]
    \normalsize
    \caption{The Design and FLOPs of each tail in MT-ViT.}
    \centering
    \setlength{\tabcolsep}{1.8mm}{
    \begin{tabular}{|c|c|c|c|c|}
   \hline\rowcolor{mygray}
         & & \multicolumn{3}{c|}{\textbf{FLOPs(G)}}  \\ \cline{3-5}\rowcolor{mygray}
         \raisebox{1.3ex}[1.3ex]{\textbf{Backbone}}&\raisebox{1.3ex}[1.3ex]{\textbf{Patch Embedding}}& \textbf{ST} & \textbf{MT} & \textbf{LT} \\ \hline\hline
         T2T-ViT-7 & T2T-module & 0.3 &0.55 &1.1 \\ 
         T2T-ViT-12 & T2T-module & 0.33 & 0.7 & 1.78 \\ 
         DeiT-Ti & Convolution & 0.31 & 0.61 & 1.25 \\ 
         DeiT-S & Convolution & 1.14 & 2.3 & 4.6  \\ 
        DeiT-B & Convolution & 4.41 & 8.9 & 17.6 \\
    \hline
    \end{tabular}
    }
    \label{flops}
\end{table}
\subsubsection{Backbones}
We implement MT-ViT on top of two popular ViT backbones (\textit{i.e.}, DeiT-Ti/S \citep{touvron2021training} and T2T-ViT-7/12 \citep{yuan2021tokens}). The FLOPs is given in Table \ref{flops}.
As for the tail predictor, we choose the light-weight MobileNetv3-small \citep{howard2019searching} as the backbone since we want to minimize the computational influence of scale predictor to the whole framework as far as possible. The number of parameters and FLOPs of MobileNetv3-small is 2.54M and 0.06G, respectively.
Followed with DVT \citep{wang2021not}, MT-ViT employs three tails, \textit{i.e.}, Short Tail (ST), Middle Tail (MT) and Long Tail (LT), to output different numbers of tokens, \textit{i.e.}, 7$\times$7, 10$\times$10, 14$\times$14. For DeiT, a convolutional kernel with $p\times p$ size and $p$ stride can be used to create non-overlap tokens. So for a 224$\times$224 image, we can obtain 7$\times$7, 10$\times$10 and 14$\times$14 tokens by setting $p$ to 32, 23 and 16 in different tails. Notice that for the middle tail, the images should be firstly resized to 230$\times$230 resolution.

As for T2T-ViT, the T2T module is used to produce a sequence of tokens. The three soft split procedures are mainly responsible for controlling the number of tokens in T2T-ViT. In each soft split, the patch size is $p\times p$ with $s$ overlapping and $k$ padding on the image, where $p-s$ is similar to the stride in convolution operation. So for an image $X \in \mathbb{R}^{h\times w\times c}$, the number of output tokens $N$ after soft split is
\begin{align}
    N=\lfloor \frac{h+2k-p}{p-s}+1 \rfloor \times \lfloor\frac{w+2k-p}{p-s}+1 \rfloor.
\end{align}
In the long tail, the patch size for the three soft splits is $p=[7,3,3]$, and the overlapping is stride $s=[3,1,1]$, which reduces the size of the input image from 224$\times$224 to 14$\times$14. By setting $s=[5,1,1]$ and $p=[11,3,3]$, the middle tail can produce 10$\times$10 tokens. By setting $s=[7,1,1]$ and $p=[14,3,3]$, the short tail can produce 7$\times$7 tokens. Therefore, we can adopt multiple tails and obtain MT-ViT backbone.

\begin{table*}[!t]
\normalsize
    \caption{The performance of MT-ViT in CIFAR100 and TinyImageNet. $\eta$ and $\alpha$ are set to 0.25, 0.75 for MT-ViT(A*), and 1, 0.25 for MT-ViT(S*).}
    \centering
    \begin{tabular}{|c|c|c|c|c|c|}\hline
   \rowcolor{mygray}
    & & \multicolumn{2}{c|}{\textbf{CIFAR100}} & \multicolumn{2}{c|}{\textbf{TinyImageNet}} \\\cline{3-6}\rowcolor{mygray}
    \raisebox{1.3ex}[1.3ex]{\textbf{Backbone}}& \raisebox{1.3ex}[1.3ex]{\textbf{Method}}& \textbf{Top-1 Acc.(\%)} & \textbf{FLOPs.(G)}  & \textbf{Top-1 Acc.(\%)} & \textbf{FLOPs.(G)} \\
    \hline\hline
    \multirow{3}*{T2T-ViT-7}&MT-ViT(A*)&82.8(+1.1)&0.71(-35.2\%)&86.6(+1.6) &0.93(-15.9\%) \\ 
    &MT-ViT(S*)&81.8(+0.1)&0.49(-55.9\%)&85.1(+0.1)& 0.57(-48.2\%)  \\ \cline{2-6}
    &Baseline &81.7 & 1.1   &85.0 & 1.1    \\  \hline\hline
                
    \multirow{3}*{T2T-ViT-12}&MT-ViT(A*)&85.7(+2.2)&1.17(-34.2\%) &90.0(+1.5)&1.41(-21.0\%)\\ 
    &MT-ViT(S*)& 84.0(+0.5)&0.53(-70.4\%)& 88.6(+0.1)&0.85(-58.5\%)\\ \cline{2-6}
    &Baseline &83.5 & 1.8   &88.5 &1.8  \\ \hline\hline
                
    \multirow{3}*{DeiT-Ti}&MT-ViT(A*)&84.9(+2.0)&0.78(-37.2\%)  &88.8(+2.1)&0.99(-20.86\%)\\
    &MT-ViT(S*) &83.4(+0.5)&0.47(-62.0\%)&86.7(+0.0)&0.47(-64.0\%)\\ \cline{2-6}
    &Baseline &82.9 & 1.3 &  86.7 &1.3  \\ \hline\hline
                
    \multirow{3}*{DeiT-S}&MT-ViT(A*)&87.9(+0.9)&2.59(-43.6\%)&94.3(+1.4)&3.18(-31.0\%) \\ 
    &MT-ViT(S*)&87.0(+0.0)&1.2(-72.0\%)&93.5(+0.6)&2.21(-52.0\%) \\  \cline{2-6}
    &Baseline &87.0 & 4.6  &92.9 & 4.6  \\ \hline
    \end{tabular}
    \label{cifar100}
\end{table*}
\subsubsection{Metric}
The reported FLOPs considers both tail predictor and MT-ViT.
Supposing $f_{st}$, $f_{mt}$ and $f_{lt}$ are the FLOPs of three tails, the overall FLOPs is calculated by 
$\frac{(f_{st}\times n_{st}+ f_{st}\times n_{mt} +f_{st}\times n_{lt})}{(n_{st}+n_{mt}+n_{lt})}+f_{predictor}$
, where $n_{st}$, $n_{mt}$ and $n_{lt}$ are numbers of images that have been processed by each individual tail and $f_{predictor}$ is the FLOPs of the tail predictor.

\subsubsection{Implementation Details}
The whole training mainly contains two processes. We first pre-train the MT-ViT backbone and then jointly finetune both tail predictor and MT-ViT in an end-to-end fashion. In the backbone pre-training, all tails are activated. The pre-training setting is basically the same as that in the official implementation of DeiT and T2T-ViT. MT-ViT backbone is trained for 300 epochs on ImageNet-1K.
For small-scale experiments, we transfer the pre-trained MT-ViT backbone to the downstream datasets such as CIFAR100 and TinyImageNet. Following the implementation of T2T-ViT, we finetune the pre-trained MT-ViT backbone for 60 epochs by using an SGD optimizer and cosine learning rate decay.  
When it comes to the finetune step, a tail predictor is introduced to determine which tail is suitable for the image. We jointly finetune the MT-ViT backbone and the predictor for 30 epochs.

\subsection{Experiment on Small-Scale Datasets}
We first conduct small-scale experiments on CIFAR100 and TinyImageNet. We implement MT-ViT on four versions of DeiT/T2T-ViT backbones, \textit{i.e.}, DeiT-Ti, DeiT-S, T2T-ViT-7 and T2T-ViT-12. The experimental results are shown in Table \ref{cifar100}. MT-ViT(A*) and MT-ViT(S*) are the same methods but they use different $\eta$ and $\alpha$ to adjust the trade-off between accuracy and speed. The former aims to achieve higher accuracy while the latter pursues a lower computational cost.

\begin{table*}[!tbp]
    \centering
    \caption{Main results of MT-ViT and other compared methods with three different ViT backbones on ImageNet-1K benchmark.}
    \setlength{\tabcolsep}{0.6mm}{
    \begin{tabular}{|c|l|c|c|c|c|}\hline
    \rowcolor{mygray}
        \textbf{Backbone}& \textbf{Method} & \textbf{Top-1 Acc.(\%)} & \textbf{Top-5 Acc.(\%)} & \textbf{FLOPs(G)} & \textbf{FLOPs$\downarrow$(\%)} \\
      \hline
      \hline
      \multirow{10}*{DeiT-Ti}  & Baseline & 72.2 & 91.1 & 1.3 & 0 \\
        & SCOP \citep{tang2020scop} (NeurIPS, 2020) & 68.9(-3.3)& 89.0(-2.1) & 0.8 & -38.5 \\
        & PoWER \citep{goyal2020power} (ICML, 2020) & 69.4(-2.8)& 89.2(-1.9) & 0.8 & -38.5 \\
        & HVT-Ti \citep{pan2021scalable} (ICCV, 2021) & 69.6(-2.6)& 89.4(-1.7) & 0.7 & -46.2 \\
        & SViTE \citep{chen2021chasing} (NeurIPS, 2021) & 71.8(-0.4)& 90.8(-0.3) & 1.0 & -23.1 \\
        & DynamicViT-$\rho$/0.7 \citep{rao2021dynamicvit} (NeurIPS, 2021) & 71.0(-1.2)& 90.4(-0.7) & 0.8 & -38.5 \\
        & DynamicViT-$\rho$/0.9 \citep{rao2021dynamicvit} (NeurIPS, 2021) & 72.3(+0.1)& 91.2(+0.1) & 1.0 & -23.1 \\
        & Evo-ViT \citep{xu2022evo} (AAAI, 2022) & 72.0(-0.2) & 91.0(-0.1)  & 0.8 & -38.5 \\
        & PS-ViT \citep{tang2022patch} (CVPR, 2022) & 72.0(-0.2)& 91.0(-0.1) & 0.7 & -46.2 \\
        & A-ViT \citep{yin2022vit} (CVPR, 2022) & 71.0(-0.2) & 90.4(-0.7)  & 0.8 & -38.5 \\
        \cline{2-6}
        & MT-ViT (Ours) & \textbf{72.9(+0.7)}& \textbf{91.3(+0.2)} & 0.8 & -38.5 \\  \hline
      
       \multirow{13}*{DeiT-S}       & Baseline & 79.8 & 95.0 & 4.6 & 0 \\
        & SCOP \citep{tang2020scop} (NeurIPS, 2020) & 77.5(-2.3)& 93.5(-1.5) & 2.6 & -43.6 \\
        & PoWER \citep{goyal2020power} (ICML, 2020) & 78.3(-1.5)& 94.0(-1.0) & 2.7 & -41.3 \\
        & HVT-S \citep{pan2021scalable} (ICCV, 2021) & 78.0(-1.8)& 93.8(-1.2) & 2.4 & -47.8 \\
        & SViTE \citep{chen2021chasing} (NeurIPS, 2021) & 79.2(-0.6)& 94.5(-0.5) & 3.0 & -34.8 \\
        & IA-RED$^2$ \citep{pan2021ia} (NeurIPS, 2021) & 79.1(-0.7)& 94.5(-0.5) & 3.2 & -30.4 \\        
        & DynamicViT-$\rho$/0.7 \citep{rao2021dynamicvit} (NeurIPS, 2021) & 79.3(-0.5)& 94.7(-0.3) & 2.9 & -37.0 \\
        & DynamicViT-$\rho$/0.9 \citep{rao2021dynamicvit} (NeurIPS, 2021) & 79.8(-0.0)&  \textbf{94.9(-0.1)}  & 4.0 & -13.0 \\
        & Evo-ViT \citep{xu2022evo} (AAAI, 2022) & 79.4(-0.4) &94.8(-0.2)  & 3.0 & -34.8  \\
        & PS-ViT \citep{tang2022patch} (CVPR, 2022) & 79.4(-0.4)& 94.7(-0.3) & 2.6 & -43.5 \\
        & EViT \citep{xu2022evo} (ICLR, 2022) & 79.5(-0.3) & 94.8(-0.2)   & 3.0 & -34.8  \\
        & ATS \citep{fayyaz2022adaptive} (ECCV, 2022) & 79.7(-0.2) & 94.9(-0.1) & 2.9 & -37.0  \\        
 	\cline{2-6}
        & MT-ViT(S*) (Ours) & 79.5(-0.3)& 94.4(-0.6) & 2.5 & -45.7 \\
        & MT-ViT(A*) (Ours) &  \textbf{80.3(+0.5)}&  \textbf{94.9(-0.1)} & 3.5 & -23.9 \\ 
        \hline
        \multirow{6}*{DeiT-B}  & Baseline & 81.8 & 95.6 & 17.6 & 0\\
        &SCOP \citep{tang2020scop} (NeurIPS, 2020)& 79.7(-2.1) & 94.5(-1.1) & 10.2 & -42.0\\ 
        &PoWER \citep{goyal2020power} (ICML, 2020)& 80.1(-1.7) & 94.6(-1.0) & 10.4 & -39.2\\
        & DynamicViT-$\rho$/0.7 \citep{rao2021dynamicvit} (NeurIPS, 2021) & 81.3(-0.5)& 95.3(-0.3) & 11.5 & -35.0 \\
        & DynamicViT-$\rho$/0.9 \citep{rao2021dynamicvit} (NeurIPS, 2021) & \textbf{81.8(-0.0)} & 95.5(-0.1)  & 15.5 & -14.0 \\
        \cline{2-6}
        &MT-ViT(Ours) & \textbf{81.8(-0.0)} & \textbf{95.6(-0.0)} & 13.9 & -21.0 \\
        \hline

    \multirow{6}*{T2T-ViT-12}& Baseline & 76.5 & 93.5 & 1.8 & 0 \\
        & PoWER \citep{goyal2020power} (ICML, 2020)& 74.5(-2.0)&92.6(-0.9) & 1.2 & -32.6  \\
        & DynamicViT-$\rho$/0.7 \citep{rao2021dynamicvit} (NeurIPS, 2021) & 76.1(-0.4)& 93.1(-0.4) & 1.2 & -32.6 \\
    & DynamicViT-$\rho$/0.9 \citep{rao2021dynamicvit} (NeurIPS, 2021) & 76.8(+0.3)& 93.5(-0.0) & 1.5 & -16.7 \\
        \cline{2-6}
        & MT-ViT (Ours) &  \textbf{77.2(+0.7)} &  \textbf{93.7(+0.2)} & 1.5 & -16.7 \\  \hline
    \end{tabular}}
    \label{sota}
\end{table*}

Compared to the baseline, it can be clearly observed that MT-ViT(A*) can outperform the baseline by 1\%-2\% while still keeping a visible advantage on FLOPs in both CIFAR100 and TinyImageNet datasets. 
For example, MT-ViT(A*) achieves 82.8\%, 85.7\%, 84.9\%, and 87.9\% on CIFAR100 with four ViT backbones, which gains an improvement of 1.1\%, 2.2\%, 2.0\% and 0.9\% over baseline, respectively. With this clear advantage in accuracy, we can still witness a decline in FLOPs of MT-ViT(A*), which is 35.2\%, 34.2\%, 37.2\%, and 43.6\%. The same thing happens in TinyImageNet. MT-ViT(A*) has roughly the same improvement in accuracy while retaining a 15\%-30\% reduction in computational cost with four ViT backbones.
By contrast, MT-ViT(S*) retains a similar accuracy with baseline but it has a significant reduction in FLOPs. For instance, MT-ViT(S*) implemented on top of DeiT-S has a similar accuracy of 87.0\% in CIFAR100 datasets, however, there is a significant decline (over 70\%) in the computational cost compared to the baseline.

\subsection{Experiment on Large-Scale Datasets}
In this subsection, we investigate the performance of MT-ViT on the large-scale benchmark ImageNet-1K. 

\subsubsection{Compared Methods}
We compare our method with several state-of-the-art methods including model pruning methods and ViT-based methods that consider exploiting multi-scale features. SCOP \citep{tang2020scop} is the state-of-the-art method in pruning the channel of CNNs. Inspired by this, we re-implement it to reduce the patches in vision transformers. PoWER \citep{goyal2020power} accelerates BERT's inference by progressively pruning tokens in BERT's layer. We directly transfer PoWER from BERT to vision transformer. DynamicViT \citep{rao2021dynamicvit}, PS-ViT \citep{tang2022patch}, SViTE \citep{pan2021ia}, Evo-ViT \citep{xu2022evo}, IA-RED$^2$ \citep{pan2021ia} and EViT \citep{liang2021evit} are six highly related compared methods that consider accelerating the inference by progressively pruning tokens of the middle layer with self-slimming or a gate function. 

\begin{figure*}[!t]
	\centering
	\includegraphics[width=2\columnwidth]{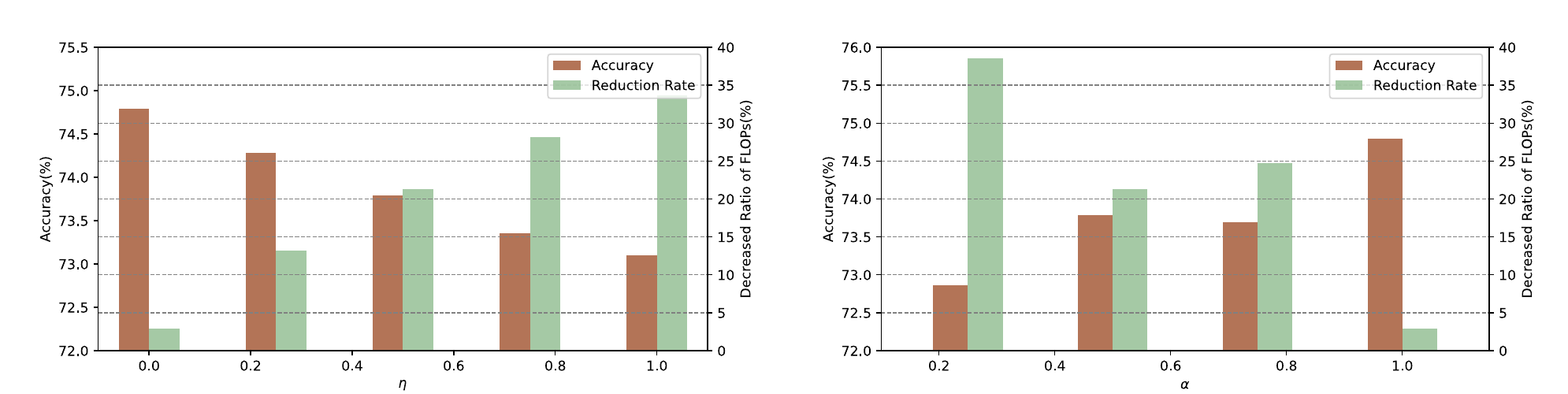}
	\caption{Ablation study on the hyper-parameter $\alpha$ and $\eta$ in FLOPs regularization term. $\alpha$ is fixed to 0.5 in the left figure and $\eta$ is fixed to 0.5 in the right figure.}
	\label{eta}
\end{figure*}
\begin{table}[!t]
    \centering
    \caption{Performance of MT-ViT when equipped with DynamicViT.}
    \setlength{\tabcolsep}{2.8mm}{
    \begin{tabular}{|l|l|l|l|}\hline
   \rowcolor{mygray}
    \textbf{Method} &  \textbf{Top-1 Acc.} & \textbf{Top-5 Acc.} & \textbf{FLOPs} \\ \hline\hline
    DynamicViT &  79.3\% & 94.7\% & 2.9G \\\hline
    MT-DynamicViT  & 79.2\% & 94.5\% & 2.4G \\
       \hline
    \end{tabular}
    }
    \label{mtdynamic}
\end{table}
\begin{table*}[!t]
    \centering
    \caption{Throughput of MT-ViT and compared methods on ImageNet-1K validation set.}
    \begin{tabular}{|l|c|c|c|}
    \hline
   \rowcolor{mygray}
        \textbf{Method} & \textbf{Top-1 Acc.(\%)} & \textbf{FLOPs(G)} & \textbf{Throughput(img/s)}  \\\hline\hline
        Baseline(DeiT-S)  & 79.8 & 4.6 & 795.53\\
        MT-ViT (Ours) & \textbf{79.5(-0.3)} & 2.5(-45.7\%) & \textbf{1677.22}\\ \hline
        \multicolumn{4}{|l|}{\textbf{Compare with Dynamic methods}} \\ \hline
        DVT \citep{wang2021not} & 79.3(-0.5) & 2.4(-47.8\%) & 1647.47\\
        DynamicViT-$\rho$-0.7 \citep{rao2021dynamicvit}  & 79.3(-0.5) & 2.9(-37.0\%) & 1222.21\\
        Evo-ViT \citep{xu2022evo}  & 79.4(-0.4) & 3.0(-34.8\%) & 1220.08\\
        E-ViT \citep{liang2021evit}  & 79.5(-0.3) & 3.0(-34.8\%) & 1219.68\\ 
        \hline
    \end{tabular}
    \label{throughput}
\end{table*}
\begin{table*}[!t]
    \centering
    \caption{Performance of MT-ViT when setting different numbers of tokens for different tails.}
    \begin{tabular}{|c|c|c|c|c|c|c|}\hline
   \rowcolor{mygray}
    \textbf{Method} & \textbf{ST} & \textbf{MT} & \textbf{LT} & \textbf{Top-1 Acc.(\%)} & \textbf{Top-5 Acc.(\%)} & \textbf{FLOPs(G)} \\ \hline\hline
    \multirow{2}*{MT-ViT} & 7$\times$7  & 10$\times$10 &  14$\times$14 & \textbf{72.9(+0.7)} & \textbf{91.3(+0.2)} & \textbf{0.8(-38.5\%)} \\ 
        & 4$\times$4& 7$\times$7& 14$\times$14&72.4(+0.2) & 91.1(+0.0) & 0.9(-30.8\%) \\ \hline
    DeiT-Ti & \multicolumn{3}{c|}{N/A} & 72.2 & 91.1&1.3 \\
        \hline
    \end{tabular}
    \label{tokens}
\end{table*}
\subsubsection{Performance}
The results are shown in Table \ref{sota}. Since SCOP \citep{tang2020scop} is designed for the pruning in the CNNs model, it is not surprising that SCOP does not perform quite well in this migration. Though there is an approximating 40\% reduction on the FLOPs, its accuracy drops rapidly as well (\textit{i.e.}, -3.3\% in DeiT-Ti and -2.3\% in DeiT-S).
PoWER \citep{goyal2020power} performs only slightly better than SCOP, which suggests that the Transformer pruning in NLP fields is not the optimal solution for the vision transformer. By contrast, the pruning methods designed for vision transformer, \textit{i.e.}, PS-ViT \citep{tang2022patch}, SViTE \citep{pan2021ia}, DynamicViT \citep{rao2021dynamicvit}, Evo-ViT \citep{xu2022evo}, IA-RED$^2$\citep{pan2021ia} and EViT \citep{liang2021evit}, are significantly better. For example, all these three methods can decrease the FLOPs up to $40\%$ in DeiT-S, with a small reduction of accuracy (less than 0.7\%). 

Our MT-ViT performs even better than these two methods, which achieves a similar FLOPs reduction but reaches higher accuracy. 
For example, with DeiT-Ti model, a nearly 40\% reduction in FLOPs makes MT-ViT comparable with other comparison methods in the computational cost. However, MT-ViT can outperform the baseline by 0.7\% in Top-1 accuracy and 0.2\% in Top-5 accuracy while other competing methods are clearly below the baseline. When running in a high-speed mode with the DeiT-S model, the accuracy and FLOPs of MT-ViT are comparable with other methods. When it comes to the high-accuracy mode, MT-ViT can outperform competing methods at least 0.5\%. 

We also investigate the performance of MT-ViT when using pruning methods as backbone. We choose one of the most representative pruning methods DynamicViT here. Based on the pre-trained Multi-Tailed Vision Transformer backbone, we employ 3 groups of predictors for 3 tails, and each group contains 3 predictors in the 3rd, 6th, and 9th layer of transformer respectively. During the pre-training, we jointly optimize the corresponding predictors and transformer in each tail for 10 epochs, which is in total 30 epochs. After that, we incorporate the tail predictor and train MT-ViT for 15 epochs. The results are shown in Table \ref{mtdynamic} and we can observe that after applying multi-tailed approach, DynamicViT could achieve further FLOPs reduction without losing too much accuracy.

\subsubsection{Influence of $\alpha$ and $\eta$}
We also investigate how the hyper-parameter $\eta$ and $\alpha$ in FLOPs constraint regularization would influence the performance of MT-ViT. The results are shown in the Figure \ref{eta}. In the left figure, we maintain a fixed value of $\alpha$ (0.5) and vary the values of $\eta$ from 0 to 1. Conversely, in the right figure, we keep $\eta$ fixed at 0.5 and adjust $\alpha$ in the range of 0.25 to 1. The FLOPs regularization can punish the situation when the predictor selects the tail with high computational cost. Therefore, when $\lambda$ is large, it has more impact on the total loss $\mathcal{L}_{total}$, and the learned predictor tends to let more images to go through short tail, which results in low accuracy and low FLOPs. When $\alpha$ is small, it will increase $\mathcal{L}_f$ and lead to an decreasing in both accuracy and FLOPs. We can clearly observe this trend in Figure \ref{eta}. When setting a larger value for $\eta$ and a smaller value for $\alpha$, we generally observe a decrease in accuracy accompanied by an increase in FLOPs reduction. By adjusting different $\eta$ and $\alpha$, we can achieve different trade-offs for MT-ViT.

\subsubsection{Throughput}
We choose several compared methods that have officially released models and compare their throughput on the ImageNet-1K validation set. The results are reported in the following Table \ref{throughput}. 
In our experiment, we report the dataset-wise throughput, which is measured on the ImageNet-1K validation set. The throughput is evaluated on a single 2080Ti with 512 batch size. From the table, we can observe that the throughput of MT-ViT is the best among all compared methods with no degradation in accuracy.

\subsubsection{Number of Tokens for Different Tails}
Following DVT's setting \citep{wang2021not}, the three tails in MT-ViT are set to output 7$\times$7, 10$\times$10 and 14$\times$14 tokens, respectively. However, the optimal number of tokens for each tail still remains to be explored. As a result, we further investigate how the number of tokens of each tail could influence the performance of MT-ViT. Specifically, a new MT-ViT backbone (based on Deit-Ti) with 4$\times$4, 7$\times$7 and 14$\times$14 tokens is pre-trained to conduct the experiment. The result is provided in Table \ref{tokens}. MT-ViT backbone with 4$\times$4, 7$\times$7 and 14$\times$14 tokens has a lower computational cost and a relatively lower accuracy. We also observe that MT-ViT with fewer tokens is inferior to MT-ViT with more tokens in both accuracy and FLOPs after the fine-tuning.

\subsubsection{Compare with DVT and MiniViT}
DVT is a great method for efficient inference in vision transformer. By adjusting the confident threshold of output logits, the trade-off of DVT between accuracy and FLOPs can be flexible. 
To provide a thorough comparison with DVT, we draw a FLOPs-Accuracy curve to compare the performance of both methods, which are based on DeiT-S.

\begin{figure}[!t]
    \centering
    \vskip 0.05in
    \includegraphics[width=0.48\textwidth]{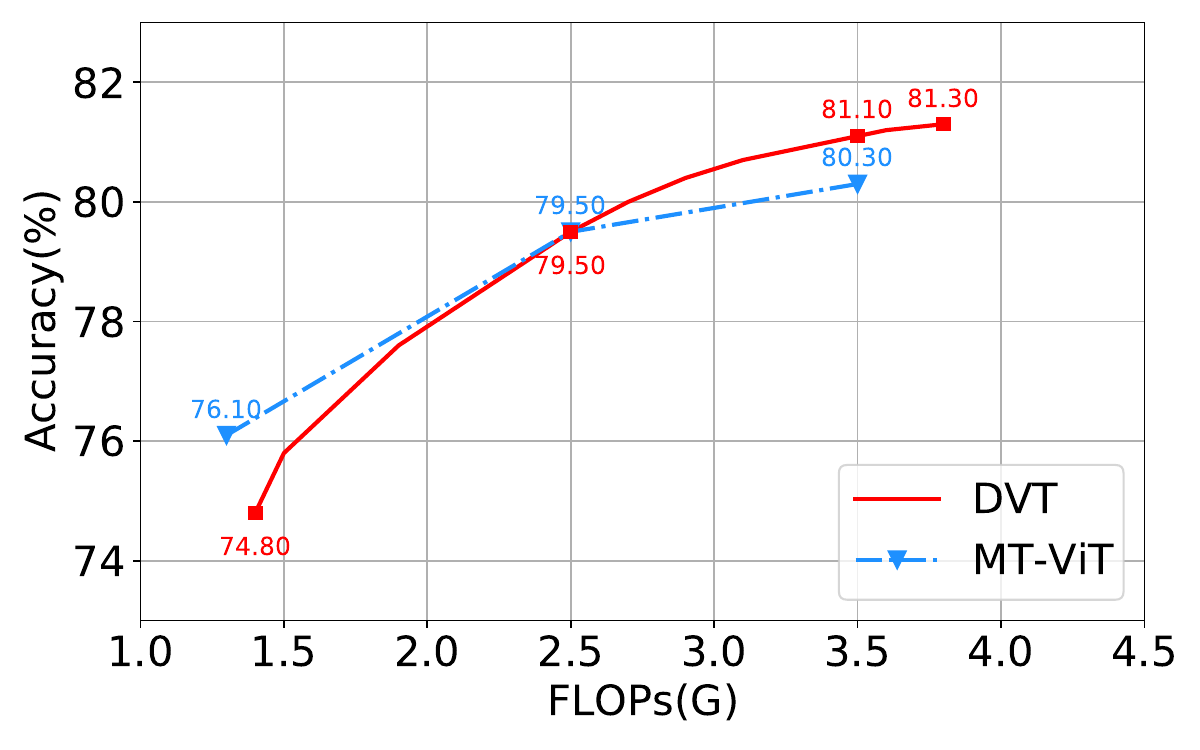}
    \caption{The Accuracy-FLOPs curve of DVT and MT-ViT based on DeiT-S.}
    \label{fig_dvt}
\end{figure}

From Figure \ref{fig_dvt}, we can observe that when running in a relatively lower FLOPs mode (less than 2.5G), MT-ViT achieves a higher accuracy than DVT. When running in a high FLOPs mode, the accuracy of DVT tends to be higher than MT-ViT. This could be attributed to the large number of parameters in DVT, which enables a stronger backbone after pre-training. However, obtaining such a backbone should also require additional computational costs. First, the number of parameters in DVT is much larger than MT-ViT. For DeiT-S, the number of DVT's parameters is 70.4M, which is around 3 times more than that of the vanilla DeiT-S (22M) and that of MT-ViT (24.5M). This could require large memory during the training and inference. Second, the training cost of DVT is much higher than MT-ViT. The training speed for DVT and MT-ViT is 751.86 img/s and 1864.87 img/s respectively. Since both methods need to pre-train the backbone for 300 epochs, it is clear that DVT requires significantly more computational resources. 
Based on the results and analysis above, we think MT-ViT is a better choice than DVT in practice, especially when the computation resources are limited.
\begin{table}[!t]
    \centering
    \caption{Comparison results of DVT, MiniViT and MT-ViT, which are implemented on top of DeiT-S. We provide number of parameters, Top-1 accuracy, and FLOPs of all three methods.}
    \setlength{\tabcolsep}{0.9mm}{
    \begin{tabular}{|l|c|c|c|}
    \hline
   \rowcolor{mygray}
        \textbf{Method} & \textbf{Param.(M)} & \textbf{Accuracy.(\%)} & \textbf{FLOPs(G)}  \\\hline\hline
        Baseline (DeiT-S)  & 22M & 79.8\% & 4.6G\\ \hline
        DVT  & 70.4M & 79.5\% (-0.3\%) & 2.5G \\
        MT-ViT(S$^\ast$) & \textbf{22M+2.5M} & 79.5\% (-0.3\%) & 2.5G \\ \hline\hline
        Mini-DeiT-S & \textbf{11M} & 80.0\% (+0.2\%) & 4.6G \\ 
        MT-ViT(A$^\ast$) & 22M+2.5M & \textbf{80.3\% (+0.5\%)} & \textbf{3.5G}\\ \hline
    \end{tabular}
    }
    \label{dvt}
\end{table}
MiniViT \citep{zhang2022minivit} is a great ViT method, which can significantly reduce the model size. From Table \ref{dvt}, we can observe that MiniViT can reduce the parameter size of the model by 50\% and there is no reduction in the computational cost. This is due to the design of MiniViT, which conducts weight sharing strategy to re-use some parameters. But this does not change either the network size or the length of the input sequence. Hence, it cannot accelerate the model for efficient inference. By contrast, the advantage of MT-ViT is that it can greatly reduce the computational cost of ViT. The additional parameters introduced by the predictor and tails are not significant.

\subsubsection{Predictor backbone}
MobileNet-v3-Small~\citep{howard2019searching} is served as the backbone for the tail predictor in the paper due to its low computational cost. However, the lower computational cost may also lead to a worse representation ability and wrong prediction. To find out how the network scale could influence the performance of the tail predictor, we use a relatively large backbone (\textit{i.e.}, ResNet with four basic blocks). The result in Table \ref{predictor} shows that improving the network scale of the predictor can only slightly improve its prediction, however, it leads to a clear growth on FLOPs. $\ast$FLOPs(G) denotes the summed FLOPs of both the tail predictor and multi-tailed vision transformer backbone. Considering the trade-off between accuracy and FLOPs, we choose to use MobileNet as the backbone of the tail predictor.
\begin{figure*}[!t]
	\centering
	\includegraphics[width=0.99\textwidth]{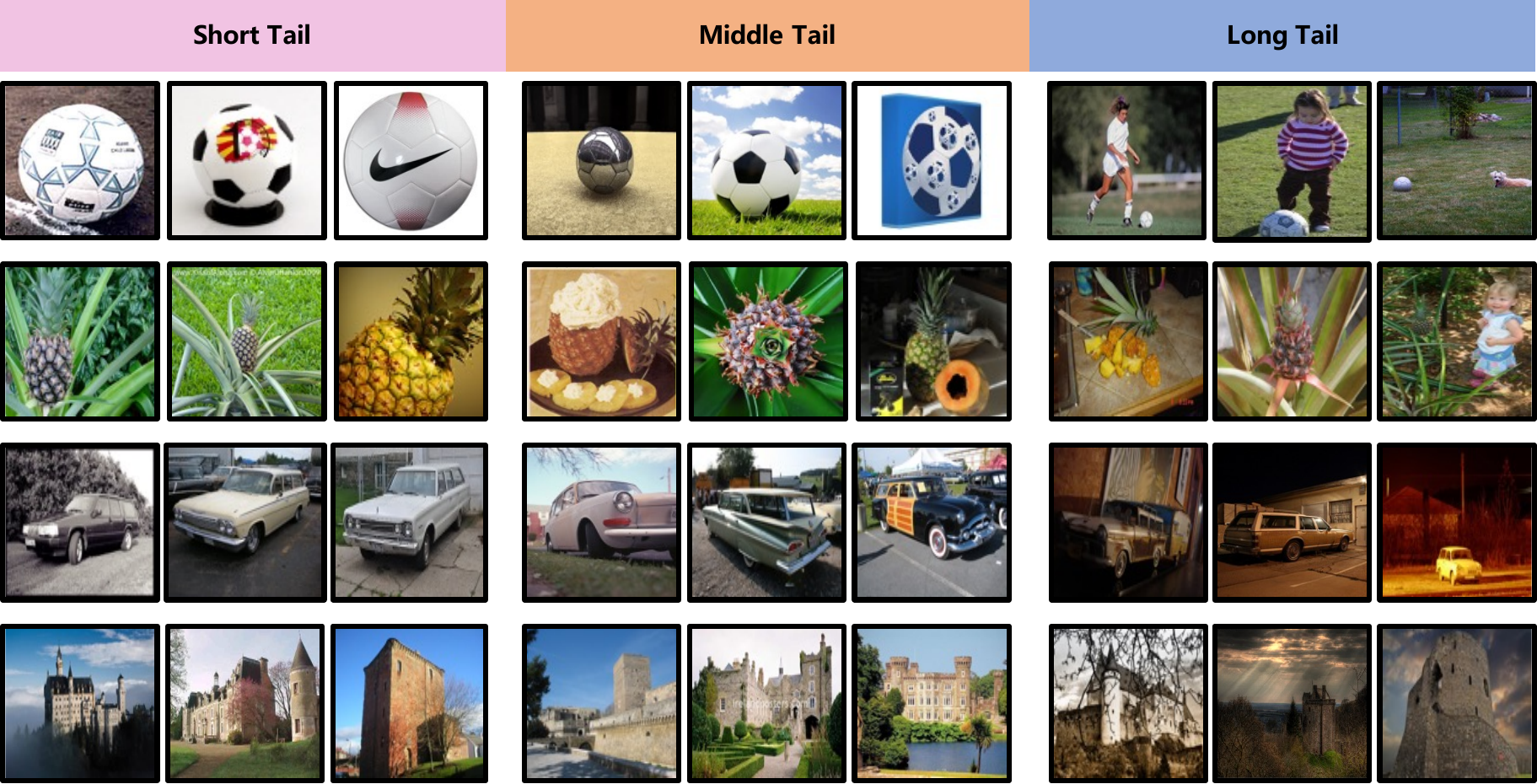}
	\caption{The visualized results of tail predictor in ImageNet-1K. The images in each row are from the class `Soccer', `Pineapple', `Car' and `Castle', respectively. The decision of each tail predictor illustrates how the predictor translates to instance difficulty.}
	\label{predictor1}
\end{figure*}
\begin{table}[!t]
    \centering
    \caption{Comparison results of different predictor backbones. The resolution is the input size of the tail predictor. The results are based on MT-ViT (DeiT-S).}
    \setlength{\tabcolsep}{1.0mm}{
    \begin{tabular}{|l|c|c|c|c|} 
    \hline
   \rowcolor{mygray}
    \textbf{Backbone} & \textbf{Resolution}  & \textbf{FLOPs(G)} &  \textbf{Acc.(\%)} & \textbf{$\ast$FLOPs(G)}   \\\hline\hline
        MobileNetv3 & 224$\times224$ & 0.06 & 80.3 & 3.5 \\ 
        ResNet-10 & 112$\times112$  & 0.24  & 80.1 & 3.5 \\ 
        ResNet-10 & 224$\times224$  & 0.89  & 80.4 & 4.3   \\
        \hline
    \end{tabular}
    }
    \label{predictor}
\end{table}

\subsubsection{Visualization}
To have a better understanding of the role of the tail predictor, we further visualize the decision of the tail predictor on ImageNet-1K, as shown in Figure \ref{predictor1}. If an image can be accurately classified by the ViT model, its cross-entropy loss $\mathcal{L}_{cls}$ will be negligible. Otherwise, its cross-entropy loss will be relatively large. An `easy’ image can be accurately classified by both three tails, which means no matter which tail the predictor chooses, the term $\mathcal{L}_{cls}$ will be negligible. Under such a case, optimizing $\mathcal{L}_{total}$ will only make $\mathcal{L}_{cls}$ drops, which encourages the predictor to choose short tail for ‘easy’ image. On the other hand, for an ‘difficult’ image that can only be accurately classified by long tails, choosing short or middle tail will result in a large $\mathcal{L}_{cls}$. Therefore, minimizing $\mathcal{L}_{total}$ will take minimizing $\mathcal{L}_{cls}$ as a priority. Under such a case, choosing long tail will results in smaller $\mathcal{L}_{cls}$, and the predictor will be encouraged to choose ‘difficult’ image.

To verify this idea, we use the trained tail predictor in MT-ViT(A*) (with DeiT-S as the backbone) to make the prediction. The samples are chosen from ImageNet-1K with four categories `Soccer', `Pineapple', `Car', and `Castle' to illustrate how the tail predictor translates instance difficulty.
Intuitively, an image with clear and large objects could be identified as an `easy' image, which can also be related to a relatively large prediction confidence of the model. An `easy' image also means that it is simple to be correctly recognized by humans. We hypothesize that decisions of the tail predictor can basically follow the human's visual judgment. From Figure \ref{predictor1}, we can easily observe that the selections of short tail and middle tail are relatively easy to identify since they often have a single frontal view object located in the center of the images. However, the objects in `difficult' images selected by long tail are often blurry and in irregular shape. This confirms our motivation that the tail predictor can measure the instance difficulty. The “sorting” into easy or hard images falls out automatically, which is learned by MT-ViT.

\subsection{Experiments on Object Detection}
We conducted additional experiments using the COCO 2017 dataset~\citep{lin2014microsoft} and applied MT-ViT to the object detection task, which enabled us to test our method's generalizability ability. 

The COCO dataset is a comprehensive collection of image data that facilitates the identification, segmentation, key-point detection, and captioning of objects. This dataset comprises 328,000 images, containing 250,000 person instances labeled with keypoints, bounding boxes, and per-instance segmentation masks. The dataset also features 91 object categories, and detecting many of these objects is highly reliant on contextual information.

To incorporate our proposed method into the detection task, we use multi-tailed vision transformer backbone as the feature extractor, while Mask R-CNN~\citep{he2017mask} is used as the detector. To generate multi-scale features for the following Mask R-CNN detector, we follow the idea of ~\citep{li2022exploring} and use only the feature map of the last transformer layer to generate multi-scale features. 
In object detection task, we only adopt two different tails (i.e., ST and LT) in MT-ViT. The size of the image is set to 512$\times$512. The patch size is set to be 32 and 16 in ST and LT, which results in 256 and 1024 tokens respectively. Following common practice, we use the multi-tailed vision transformer that is pre-trained on the ImageNet-1K classification task to initialize the backbone. We train the model for 25 epochs on 4 RTX 4090 GPUs with a batch size of 32. Our implementation and training hyper-parameters are based on this repo\footnote{https://github.com/ViTAE-Transformer/ViTDet}. The FLOPs of both ST and MT in detection is measured by the official tool in mm-detection\footnote{https://github.com/open-mmlab/mmdetection/}. The results are shown in the following Table \ref{tab:objectdetection}. 
\begin{table}[!t]
\normalsize
    \centering
    \caption{The performance on COCO object detection task using Mask R-CNN. We train the models for 25 epochs with a batch size of 32. FLOPs are computed on $512\times512$ images.}
    \setlength{\tabcolsep}{0.25mm}{
    \begin{tabular}{|l|c|c|c|c|c|} \hline\rowcolor{mygray}
    &\multicolumn{2}{c}{\textbf{FLOPs}}&\multicolumn{3}{c|}{\textbf{Metric}}  
    \\ \cline{2-6}\rowcolor{mygray}
    \raisebox{1.3ex}[1.3ex]{\textbf{Backbone}} & \textbf{Overall} & \textbf{Backbone} & $mAP$ & $mAP^{box}_{50}$ & $mAP^{box}_{75}$ \\ \hline \hline
    DeiT-Ti & 46.6 & 6.5 & 32.1 & 51.8 & 33.9 \cr
    MT-ViT & 39.4\textbf{(-15.5\%)} & 5.5 & 31.9 & 50.1 & 33.8 \cr
    \hline
    \end{tabular}
    }
    \label{tab:objectdetection}
\end{table}

For the object detection task, we observe that using MT-ViT can reduce the FLOPs by 15.5\% with only slight degradation on the $mAP^{box}$(-0.2). These results of our experiments on COCO 2017 and dense prediction tasks provide further evidence of the robustness and versatility of our approach, demonstrating its ability to generalize across diverse datasets and tasks.

\section{Conclusion}
In this paper, we propose an efficient vision transformer called Multi-Tailed Vision Transformer (MT-ViT) by reducing the number of tokens in the tail of the vision transformer. MT-ViT adopts multiple tails for splitting images into sequences with varying lengths, each of which results in different computational costs and accuracy during inference. We conditionally send images to different tails by introducing a tail predictor that determines which tail is appropriate for the image. During training, the Gumbel-Softmax trick ensures that both modules can be optimized in an end-to-end fashion. The empirical results demonstrate that MT-ViT outperforms baseline and other comparison methods on small-scale datasets (\textit{i.e.}, CIFAR100 and TinyImageNet) and large-scale datasets (\textit{i.e.}, ImageNet-1K). The visualized results of the tail predictor also identify our motivation that the tail predictor can be employed to automatically translate instance difficulty.

\bibliographystyle{cas-model2-names}

\bibliography{mtvit1}
\balance



\end{document}